\documentclass[letterpaper]{article} % DO NOT CHANGE THIS
\usepackage{aaai23-r2hcai}  % DO NOT CHANGE THIS
\usepackage{times}  % DO NOT CHANGE THIS
\usepackage{helvet}  % DO NOT CHANGE THIS
\usepackage{courier}  % DO NOT CHANGE THIS
\usepackage[hyphens]{url}  % DO NOT CHANGE THIS
\usepackage{graphicx} % DO NOT CHANGE THIS
\usepackage{natbib}  % DO NOT CHANGE THIS AND DO NOT ADD ANY OPTIONS TO IT
\usepackage{caption} % DO NOT CHANGE THIS AND DO NOT ADD ANY OPTIONS TO IT
\usepackage{algorithm}
\usepackage{algorithmic}
\usepackage{amsthm}
\usepackage{amsmath}
\usepackage{booktabs}
\usepackage{multirow}
\usepackage{multicol}
\usepackage[table]{xcolor} % Essential for row coloring

\usepackage{newfloat}
\usepackage{listings}
\DeclareCaptionStyle{ruled}{labelfont=normalfont,labelsep=colon,strut=off} % DO NOT CHANGE THIS
\floatstyle{ruled}
\newfloat{listing}{tb}{lst}{}
\floatname{listing}{Listing}
\usepackage{fancyhdr}
\fancypagestyle{firstpagehf}
\title{PAIRS: Parametric--Verified Adaptive Information Retrieval and Selection for Efficient RAG}
\author{
    Wang Chen\textsuperscript{\rm 1,2},
    Guanqiang Qi\textsuperscript{\rm 1},
    Weikang Li\textsuperscript{\rm 3},
    Yang Li\textsuperscript{\rm 1}\footnote{Corresponding author},
    Deguo Xia\textsuperscript{\rm 1},
    Jizhou Huang\textsuperscript{\rm 1}
}
\affiliations{
    \textsuperscript{\rm 1} Baidu Inc\\
    \textsuperscript{\rm 2} The University of Hong Kong\\
    \textsuperscript{\rm 3} Peking University \\
    \{chenwang05, qiguanqiang, liyang164, xiadeguo, huangjizhou01\}@baidu.com, wavejkd@pku.edu.cn
}

\begin{document}
\thispagestyle{firstpagehf}
\maketitle

\begin{abstract}
Retrieval--Augmented Generation (RAG) has become a cornerstone technique for enhancing large language models (LLMs) with external knowledge. However, current RAG systems face two critical limitations: (1) they inefficiently retrieve information for every query, including simple questions that could be resolved using the LLM's parametric knowledge alone, and (2) they risk retrieving irrelevant documents when queries contain sparse information signals. To address these gaps, we introduce \textbf{P}arametric--verified \textbf{A}daptive \textbf{I}nformation \textbf{R}etrieval and \textbf{S}election (\textbf{PAIRS}), a training-free framework that integrates parametric and retrieved knowledge to adaptively determine whether to retrieve and how to select external information. Specifically, PAIRS employs a dual-path generation mechanism: First, the LLM produces both a direct answer and a context--augmented answer using self-generated pseudo-context. When these outputs converge, PAIRS bypasses external retrieval entirely, dramatically improving the RAG system's efficiency. For divergent cases, PAIRS activates a dual-path retrieval (\textbf{DPR}) process guided by both the original query and self-generated contextual signals, followed by an Adaptive Information Selection (\textbf{AIS}) module that filters documents through weighted similarity to both sources. This simple yet effective approach can not only enhance efficiency by eliminating unnecessary retrievals but also improve accuracy through contextually guided retrieval and adaptive information selection. Experimental results on six question--answering (QA) benchmarks show that PAIRS reduces retrieval costs by around 25\% (triggering for only 75\% of queries) while still improving accuracy---achieving +1.1\% EM and +1.0\% F1 over prior baselines on average.

\end{abstract}

\section{Introduction}

Large language models (LLMs) have dramatically advanced natural language processing (NLP), achieving comparable or even better performance than human beings \cite{touvron2023llama, achiam2023gpt, guo2025deepseek, yang2025qwen3}. However, previous studies \cite{ji2023survey} found that LLMs can only answer questions or accomplish tasks by leveraging their parametric knowledge, which cannot update the latest information nor private datasets. To address this issue, retrieval--augmented generation (RAG) framework was designed to enhance the generation performance of LLMs using the retrieved information \cite{lewis2020retrieval, guu2020retrieval, karpukhin2020dense}. The effectiveness of RAG has been validated across various NLP tasks, achieving impressive improvement compared with pure LLM systems \cite{ram2023context, gao2023retrieval}.

Despite its widespread adoption, conventional RAG systems suffer from two critical shortcomings. First, they inefficiently trigger retrieval for every query, including straightforward questions that could be resolved solely using the LLM’s internal parametric knowledge, resulting in unnecessary computational latency and resource consumption. Second, they exhibit vulnerability to sparse queries---concise user inputs with limited contextual signals---which often yield irrelevant or low-quality retrievals due to inadequate semantic cues, ultimately degrading answer accuracy and reliability \cite{wang2023query2doc, gao2023precise}.

Existing approaches address these challenges through two primary strategies: query augmentation (e.g., appending LLM--generated pseudo-documents or external data to enrich sparse queries) \cite{jagerman2023query, buss2023generating, jeong2024database} and retrieval optimization (e.g., adopting reinforcement learning or rerankers to refine document retrieval or selection) \cite{asai2024selfrag, jin2025search, chang-etal-2025-main}. However, most of these methods may suffer from high computational costs and overlook LLMs' inherent capability to resolve simple queries without retrieval.

\begin{figure*}[!ht]
\centering
\includegraphics[width=1.9\columnwidth]{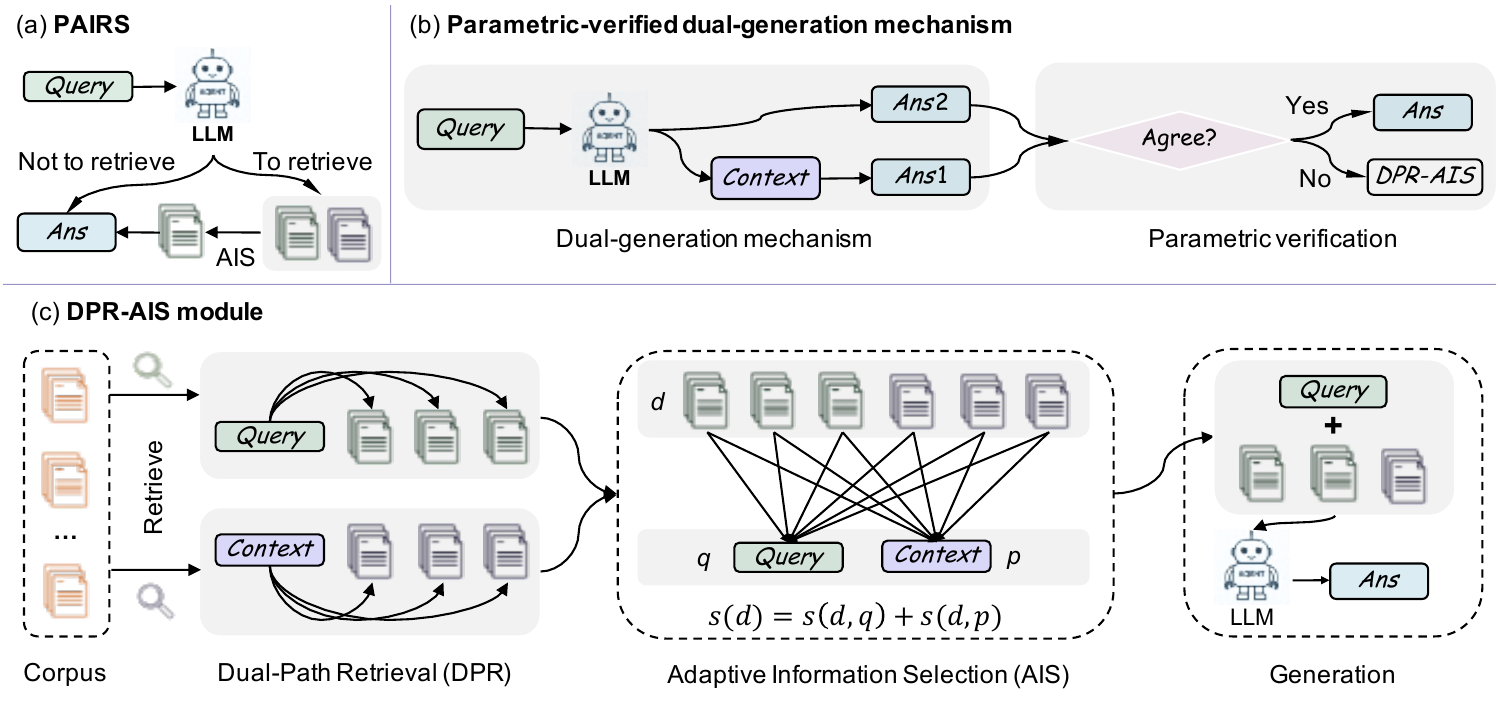}
\caption{Illustration of the PAIRS framework. (a) PAIRS can adaptively determine whether to retrieve and how to select external information. (b) PAIRS applies a dual-generation mechanism to verify the correctness of answers. (c) PAIRS adaptively selects effective information based on the dual-path retrieval for the final generation.}
\label{fig: framework}
\end{figure*}

Recently, the emerging LLMs contain billions of or even more than 1 trillion parameters and are trained on massive datasets, endowing them with rich parametric knowledge that enables them to answer many simple queries without external retrieval. To explore how this capability can enhance both the efficiency and effectiveness of RAG systems, we propose PAIRS (Parametric--verified Adaptive Information Retrieval and Selection), a \textbf{training-free}, simple yet effective framework that enhances RAG by adaptively integrating parametric and retrieved knowledge (as shown in Fig. \ref{fig: framework}). The key contributions of this work are summarized as follows:
\begin{itemize}
    \item \textbf{Parametric--verified dual-path mechanism}: PAIRS introduces a novel parametric verification approach, where the LLM generates both a direct answer and a pseudo-context--augmented answer. If the two answers agree, it confidently bypasses retrieval, significantly improving efficiency.

    \item \textbf{DPR-AIS module}: For divergent cases, we propose a Dual-Path Retrieval (DPR) strategy that retrieves documents using both the query and the generated pseudo-context. These are then filtered via the Adaptive Information Selection (AIS) module, which scores each document by weighted similarity to both sources.

    \item \textbf{Efficiency with accuracy}: Experimental results demonstrate that PAIRS can reduce reliance on external retrieval, triggering it for only approximately 75\% of queries, while still achieving higher EM (+1.1\%) and F1 (+1.0\%) scores than prior baselines across six QA datasets on average.

    \item \textbf{Flexible integration}: The framework is modular and can be seamlessly combined with other enhancements, such as reranking models, to further boost performance. For example, DPR-AIS-rerank achieves state-of-the-art results, outperforming strong reranking baselines by 2.3\% EM and 2.5\% F1 on average across six datasets.
\end{itemize}

%%%%%%%%%%%%%%%%%%%%%%%%%%%%%%% LR %%%%%%%%%%%%%%%%%%%%%%%%%%%%%%%
\section{Related work}

In this section, we review the related works in recent years and further discuss the research gaps. Specifically, we primally focus on two folds: (1) how to augment a query to enhance retrievals and (2) how to retrieve and select relevant information.

\subsection{Query augmentation}

A query from users may be concise and contain relatively sparse information, which can result in suboptimal retrievals and poor answers. To address this issue, many studies have proposed to augment the query by appending additional terms extracted from retrieved documents \cite{lavrenko2017relevance, lv2009comparative} or generated by neural models \cite{zheng2020bert, mao2021generation}. Recently, many studies leveraged advanced LLMs to augment queries and achieved notable improvement \cite{buss2023generating, wang2023query2doc, jagerman2023query, gao2023precise, lin2023train}. For example. \citet{wang2023query2doc} first prompted LLMs to generate a pseudo document of the query, which was then concatenated with the original query to enhance the retrieval. In addition, a few studies \cite{jeong2022augmenting, jeong2024database} used an external database, such as relevant tabular data, to augment original queries, which can enhance query representations but require additional data. Furthermore, a few studies \cite{chan2024rq, zhang2025imprag} trained models to refine or encode queries to enhance the performance of information retrieval and question answering. However, the aforementioned approaches did not fully leverage the parametric knowledge of LLMs that can answer a proportion of queries directly. In addition, they did not explore more effective combination methods of the query and generated context to retrieve and select more relevant information.

\subsection{Information retrieval and selection}

How to retrieve and select relevant information is a key issue in RAG systems. As the remarkable success of advancing LLMs using reinforcement learning (RL), many studies have adopted RL to train LLMs to enhance information retrieval and reranking \cite{asai2024selfrag, jin2025search, li2025search, song2025r1, yu2024rankrag}. Also, a few studies \cite{jiang2025gainrag, yan2025rpo} aligned the retriever's preferences with LLMs to retrieve more relevant information and improve the generation performance. These methods could achieve significant improvement but may suffer from high computational costs. In addition, with the accessibility to advanced reranking models, such as bge-reranker \cite{bge_embedding} and Qwen3-reranker \cite{zhang2025qwen3}, a straightforward method is to first rerank the retrievals using a reranker and then select the top-$k$ relevant retrievals \cite{glass2022re2g, chang-etal-2025-main}. This method could further select retrievals with a higher semantic similarity with the query and thus enhance the generation performance. However, it may not perform well or even degrade answer accuracy in QA tasks due to the sparse query or large corpus \cite{kimsure, jiang2025gainrag}.

%%%%%%%%%%%%%%%%%%%%%%%%%%%%%%% Method %%%%%%%%%%%%%%%%%%%%%%%%%%%%%%%
\section{Methodology}\label{sec: method}

In this section, we introduce the preliminary problem and the details of our proposed method.

\subsection{Preliminary}

In a retrieval--augmented generation (RAG) system, a collection of documents is first split into small chunks $D$, which are then embedded into vectors using an embedding model $f$. Given a query $q$, the retriever $\mathcal{R}$ first searches top $k$ relevant chunks $D_k = \{d_1, d_2, ..., d_k\} \subset D$ from the collection according to the similarity between the query and chunks, i.e., $D_k = \mathcal{R}(D, q)$. The similarity $s$ is calculated using the inner product (IP) or L2 norm between the query embedding $\mathbf{q} = f(q)$ and chunk embeddings $\mathbf{d} = f(d), \forall d \in D$. Finally, the retrievals and query are incorporated into a prompt $\mathcal{P}(q, D_k)$, which is used as the input of the generator model $g$ to generate final answer, i.e.,  $\hat{a} = \mathcal{G}(\mathcal{P}(q, D_k))$. We list all notations and abbreviations in the appendix.

\subsection{Parametric verification mechanism}

Modern LLMs \cite{achiam2023gpt, guo2025deepseek, yang2025qwen3} are equipped with powerful parametric knowledge acquired from large-scale pretraining corpora, enabling them to answer many questions without the need for external information. To exploit this capability, PAIRS incorporates a parametric verification mechanism that adaptively determines whether external retrieval is necessary. At its core is a dual-generation strategy, where the LLM produces two responses to a given query $q$: one generated directly from its parametric knowledge, i.e., $\hat{a}_1 = \mathcal{G}(\mathcal{P}(q))$, and the other based on a self-generated pseudo-context, i.e., $\hat{a}_2 = \mathcal{G}(\mathcal{P}(q, p))$. This pseudo-context $p$ mimics retrieved content but is synthesized internally by the LLM, ensuring the generation process remains efficient and self-contained. If the two responses converge--- the same or demonstrating semantic consistency---PAIRS concludes that external retrieval is unnecessary and directly returns the answer. This approach not only reduces computational overhead but also avoids the potential pitfalls of retrieving irrelevant or noisy documents. On the other hand, if the two responses diverge, suggesting uncertainty in the model’s parametric understanding, the system proceeds to initiate external document retrieval and selection. In this way, the parametric verification mechanism serves as a lightweight and effective decision gate that enhances both the efficiency and robustness of the RAG pipeline.

\subsection{Dual-path retrieval}

\begin{figure}[!ht]
\centering
\includegraphics[width=1.0\columnwidth]{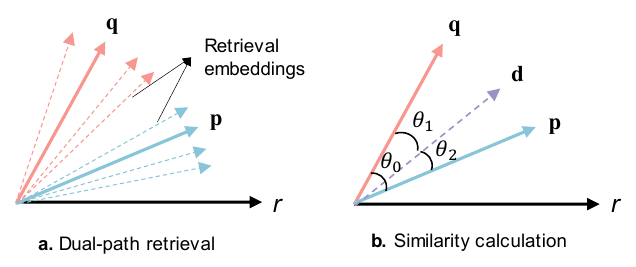}
\caption{Illustration of (a) dual-path retrieval mechanism and (b) similarity calculation of retrievals.}
\label{fig: retrieval}
\end{figure}

One of the central challenges in RAG is handling sparse or underspecified queries, which often lead to the retrieval of irrelevant or low-quality documents \cite{zhao2024retrieval, singh2025agentic}. This issue becomes particularly pronounced when relying solely on the original query to perform retrieval (please refer to the appendix for further analysis). However, the previously generated pseudo-context can be used to enhance the retrieval. An intuitive solution might be to concatenate the query with LLM--generated pseudo-context to enrich the signal \cite{wang2023query2doc, jagerman2023query}; however, such concatenation could be inefficient, as the query and pseudo-context serve different semantic roles and may not integrate cohesively. To address this, PAIRS adopts a dual-path retrieval (DPR) mechanism that treats the query and the self-generated context as complementary sources of information. Instead of combining them into a single retrieval signal, the system executes two parallel retrieval operations---one conditioned on the query and the other on the pseudo-context. Specifically, as shown in Fig. \ref{fig: retrieval} (a), the system retrieves top $n$ relevant documents using the embeddings of the query $\mathbf{q} = f(q)$ and the generated context $\mathbf{p} = f(p)$, respectively, and the retrieved documents can be denoted as $D_{2n} = \{d_1, d_2, ..., d_{2n}\} \subset D$. This dual-path strategy enhances the relevance and diversity of the retrieved documents, which is especially beneficial for complex or ambiguous queries where either source alone may fall short.

\subsection{Adaptive information selection}

Once the dual-path retrieval yields a set of $2n$ candidate documents based on the original query and the LLM-generated pseudo-context, PAIRS introduces an adaptive information selection (AIS) mechanism to refine this candidate set. The goal is to prioritize documents that are simultaneously relevant to $q$ and $p$, ensuring that the final context passed to the LLM is not only topically aligned but also semantically coherent from different perspectives. Formally, we compute the score for each document $s(d)$ as follows:
\begin{equation}
    s(d) = s_1(d, q) + s_2(d, p), \quad \forall d \in D_{2n},
\end{equation}
where $s_1(d,q)$ and $s_2(d,p)$ denote the relevance between the document $d$ and the query $q$ and LLM--generated context $p$, respectively.

As shown in Fig. \ref{tab: results} (b), a straightforward method is to calculate the relevance ($s_1$ and $s_2$) as the IP between their corresponding embeddings, as follows:
\begin{equation}
    s_1 (d, q) = \cos (\theta_1) = \langle \mathbf{d}, \mathbf{q} \rangle,
\end{equation}
\begin{equation}
    s_2(d,q)= \cos (\theta_2) = \langle \mathbf{d}, \mathbf{p} \rangle,
\end{equation}
where $\mathbf{q} = f(q)$, $\mathbf{p} = f(p)$, and $\mathbf{d} = f(d), \forall d \in D_{2n}$. Remind that $\mathbf{q}$, $\mathbf{p}$, and $\mathbf{d}$ are normalized vectors. However, such aggregation may not fully capture the joint relevance, especially if either $q$ or $p$ introduces semantic noise. Therefore, AIS leverages a geometric intuition: if both $\mathbf{q}$ and $\mathbf{p}$ align well with a given document $\mathbf{d}$, then the combined angle $\theta = \theta_1 + \theta_2$ should be minimized. This motivates the maximization of $s(d) = \cos(\theta) = \cos(\theta_1 + \theta_2)$, which expands to:
\begin{align}\label{eq: score}
    & s(d) = \cos (\theta_1 + \theta_2) \nonumber \\
    & = \cos (\theta_1) \cdot \cos (\theta_2) - \sqrt{1-(\cos (\theta_1))^2} \cdot \sqrt{1-(\cos (\theta_2))^2} \nonumber \\
    & = s_1 \cdot s_2 - \sqrt{1-(s_1)^2} \cdot \sqrt{1-(s_2)^2}.
\end{align}
This formulation implicitly rewards documents that are jointly aligned with both $\mathbf{q}$ and $\mathbf{p}$ while penalizing those that diverge from either direction. The DPR-AIS algorithm, i.e., dual-path retrieval--based adaptive information selection, is formulated as Algorithm \ref{alg: DPR-AIS}.

\begin{algorithm}[!htb]
\caption{DPR-AIS}
\label{alg: DPR-AIS}
\textbf{Input}: Query $q$, LLM--generated pseudo-context $p$,  encoder $f(\cdot)$, retriever $\mathcal{R}$, corpus $D = \{d_1, d_2, \dots, d_N\}$, retrieval size $n$, and selection size $k$\\
\textbf{Output}: Filtered relevant documents $D_k \subset D$

\begin{algorithmic}[1]
\STATE Compute embeddings: $\mathbf{q} \leftarrow f(q),\; \mathbf{p} \leftarrow f(p),\; \mathbf{d} \leftarrow f(d)$
\STATE Retrieve top-$n$ documents using $\mathbf{q}$: $D_q \leftarrow \mathcal{R}(D, \mathbf{q})$
\STATE Retrieve top-$n$ documents using $\mathbf{p}$: $D_p \leftarrow \mathcal{R}(D, \mathbf{p})$
\STATE Merge retrieved sets: $D_{2n} \leftarrow D_q \cup D_p$
\FOR{each $d \in D_{2n}$}
    \STATE Compute $s_1 = \langle \mathbf{q}, \mathbf{d} \rangle$ and $s_2 = \langle \mathbf{p}, \mathbf{d} \rangle$
    \STATE Compute joint relevance score using Eq. \eqref{eq: score}
\ENDFOR
\STATE Select top-$k$ documents from $D_{2n}$ by descending $s(d)$ values
\STATE \textbf{return} selected document set $D_k$
\end{algorithmic}
\end{algorithm}

\subsection{Extension and beyond}

The flexibility of the proposed DPR-AIS mechanism opens several promising avenues for extension and enhancement. A key observation is that the relative importance of the original query $q$ and the LLM-generated pseudo-context $p$ may vary across different scenarios. For instance, when $q$ is underspecified or ambiguous, $p$ may provide more concrete semantic signals; conversely, when $p$ is noisy or misaligned, $q$ may remain the more reliable anchor for retrieval. To accommodate this variability, we extend our selection metric by introducing a dynamic weighting scheme that modulates the contribution of each signal. Specifically, we define the weighted retrieval direction as:
\begin{equation}
\theta_{\text{dynamic}} = \alpha \cdot \theta_1 + (1 - \alpha) \cdot \theta_2,
\end{equation}
where the coefficient $\alpha \in [0,1]$ serves as an importance factor that can be determined dynamically, and a larger $\alpha$ implies greater trust in the query and less in the pseudo-context. One principled approach is to compute $\theta_0$, the angle between $\mathbf{q}$ and $\mathbf{p}$ (as shown in Fig. \ref{fig: retrieval} (b)), and use it to set $\alpha$, thereby reflecting the alignment between $q$ and $p$. Specifically, given a training set of queries and their corresponding documents, generated contexts, and their associated relevance judgments, we can empirically determine $\alpha$ for each instance as:
\begin{equation}\label{eq: alpha}
\alpha = \frac{\theta_2}{\theta_1 + \theta_2},
\end{equation}
which reflects the relative informativeness of the query $q$ compared to the generated context $p$ when retrieving a relevant document $d$. Then, we can fit a model to predict $\alpha$ from $\theta_0$ at inference time.

Furthermore, DPR-AIS can be integrated with existing reranking architectures. Rather than relying purely on embedding--based similarity, we can use the reranking scores between $q$ and candidate document $d$ as $s_1(d,q)$, and between $p$ and $d$ as $s_2(d,p)$. Therefore, the score for a document can be calculated as follows:
\begin{equation}
    s(d) = \mathcal{K}(d, q) + \mathcal{K}(d, p),
\end{equation}
where $\mathcal{K}(\cdot, \cdot)$ denotes the reranking model that evaluates the semantic relevance between two documents. This provides a novel extension to traditional reranking pipelines by considering both the explicit query and its latent expansion.

%%%%%%%%%%%%%%%%%%%%%%%%%%%%%%% Eexperiments %%%%%%%%%%%%%%%%%%%%%%%%%%%%%%%
\section{Experiments}\label{sec: exp}

In this section, we report our experiment setups, including implementation details, evaluation datasets, and baselines, main experimental results, ablation study results, and in-depth discussions.

\subsection{Experiment setup}

\textbf{Evaluations datasets.} We evaluate the effectiveness of the proposed method on open domain QA datasets: (1) NaturalQA \cite{kwiatkowski2019natural}, (2) WebQuestions \cite{berant2013semantic}, (3) SQuAD \cite{rajpurkar2016squad}, and (4) TriviaQA \cite{joshi2017triviaqa}. In addition, we further test the method using two complex multi-hop QA datasets, i.e., HotpotQA \cite{yang2018hotpotqa} and 2WikiMultiHopQA \cite{ho2020constructing}. For the open domain datasets, we use the 21M English Wikipedia dump as the corpus for retrieval, while for the multi-hop datasets, we use their original corpus. The statistics of all datasets are listed in Tab. \ref{tab: data}.

\begin{table}[!ht]
\centering
%\resizebox{.95\columnwidth}{!}{
\begin{tabular}{l l l}
\toprule
    Datasets & \#Questions & \#Documents \\
\midrule
    NaturalQA    & 3,610  & 21 M \\
    WebQuestions & 2,032  & 21 M \\
    SQuAD        & 10,570 & 21 M \\
    TriviaQA     & 11,313 & 21 M \\
\midrule
    HotpotQA        & 7,405 & 5 M \\
    2WikiMultiHopQA & 12,576 & 431 K \\
\bottomrule  
\end{tabular}
\caption{Statistics of evaluation datasets.}
\label{tab: data}
\end{table}

\begin{table*}[!ht]
\centering
\resizebox{2.1\columnwidth}{!}{
\begin{tabular}{l c c c c c c c c c c c c c c}
\toprule
    \multirow{2}{*}{Method} & \multicolumn{2}{c}{HotpotQA} & \multicolumn{2}{c}{2WikiQA} & \multicolumn{2}{c}{NaturalQA} & \multicolumn{2}{c}{WebQuestions} & \multicolumn{2}{c}{SQuAQ} & \multicolumn{2}{c}{TriviaQA} & \multicolumn{2}{c}{Average}\\
        & EM & F1 & EM & F1 & EM & F1 & EM & F1 & EM & F1 & EM & F1 & EM & F1\\
\midrule
No Retrieval & 17.27 & 23.92 & 21.55 & 24.97 & 15.54 & 18.85 & 17.67 & 24.42 & 10.54 & 16.96 & 38.40 & 20.25 & 21.16 & 21.56\\
Standard RAG & 34.84 & 44.75 & \textbf{29.19} & \textbf{34.17} & 34.29 & 36.06 & 20.08 & 27.68 & 28.99 & 34.41 & 54.69 & 30.89 & 33.68 & 34.66\\
        HyDE & 32.74 & 41.93 & 23.47 & 28.17 & 34.60 & 36.38 & 22.69 & 30.74 & 25.53 & 31.03 & 56.19 & 31.26 & 32.54 & 33.25\\
        Q2D  & 33.71 & 43.41 & 24.25 & 29.07 & 35.68 & 37.36 & \underline{22.79} & \underline{30.77} & \underline{29.20} & 34.79 & 58.12 & 32.56 & 33.96 & 34.66\\
        CoT  & 32.96 & 42.50 & 24.05 & 28.88 & 35.67 & 37.36 & 22.15 & 30.52 & 28.38 & 34.23 & 58.32 & 32.39 & 33.59 & 34.13\\
    
    \rowcolor{lightgray!30}
    PAIRS & \underline{35.14} & \underline{45.20} & 26.99 & 31.73 & \underline{36.87} & \underline{38.17} & \textbf{23.08} & \textbf{31.15} & 28.85 & \underline{34.90} & \textbf{59.23} & \underline{32.92} & \underline{35.03} & \underline{35.68}\\
    
    RA ratio & \multicolumn{2}{c}{(74.4\%)} & \multicolumn{2}{c}{(69.1\%)} & \multicolumn{2}{c}{(82.8\%)} & \multicolumn{2}{c}{(78.6\%)} & \multicolumn{2}{c}{(86.3\%)} & \multicolumn{2}{c}{(61.6\%)} & \multicolumn{2}{c}{(75.5\%)}\\
    
    \rowcolor{lightgray!30}
    DPR-AIS & \textbf{36.61} & \textbf{47.02} & \underline{28.48} & \underline{33.48} & \textbf{37.42} & \textbf{38.95} & 22.00 & 30.63 & \textbf{30.00} & \textbf{35.87} & \underline{59.09} & \textbf{33.32} & \textbf{35.60} & \textbf{36.55}\\

    \midrule
    Rerank & 38.56 & 49.48 & \textbf{31.02} & \textbf{36.04} & 33.60 & 35.12 & 20.47 & 28.83 & 26.20 & 31.48 & 57.58 & 32.64 & 34.57 & 35.60 \\
    \rowcolor{lightgray!30}
    DPR-AIS-rerank & \textbf{39.42} & \textbf{50.21} & 30.34 & 35.10 & \textbf{36.12} & \textbf{37.36} & \textbf{21.85} & \textbf{30.30} & \textbf{29.99} & \textbf{35.81} & \textbf{59.78} & \textbf{33.88} & \textbf{36.25} & \textbf{37.11}\\

\bottomrule
\end{tabular}}
\caption{Main results across six QA datasets. \textbf{RA ratio} indicates the proportion of queries for which the retriever was activated by PAIRS. \textbf{Bold} and \underline{underlined} values represent the best and second-best scores, respectively.}
\label{tab: results}
\end{table*}

\textbf{Baselines.} We compare the proposed method with other training--free baselines. (1) \textbf{No Retrieval} leverages the parametric knowledge of LLM to directly generate response to the given query without retrieval. (2) \textbf{Standard RAG} incorporates the retrieved top-$k$ documents with the query as the input of LLM through prompting. (3) \textbf{HyDE} \cite{gao2023precise} leverages the LLM--generated passage to enhance the retrieval. (4) \textbf{Q2D} \cite{wang2023query2doc} concatenates multiple queries and the LLM--generated pseudo-document to perform the retrieval. (5) \textbf{CoT} \cite{jagerman2023query} first prompts the LLM to generate the answer as well as the rationale to the given query, and then combines multiple queries and the LLM outputs into the retrieval signal. In addition, we also compare DPR-AIS--powered reranking (denoted as \textbf{DPR-AIS-rerank}) with the traditional reranking mechanism. Hence, we have the last baseline: (6) \textbf{Rerank} \cite{glass2022re2g, chang-etal-2025-main} selects the top-$k$ documents from retrievals using a reranker.

\textbf{Evaluation metrics.} To comprehensively assess the quality of generated answers, we adopt two widely used metrics in open-domain question answering: Exact Match (EM) and F1 score. The EM metric quantifies the proportion of predictions that exactly match any one of the ground-truth answers, serving as a strict measure of correctness. On the contrary, the F1 score provides a more forgiving evaluation by computing the token-level overlap between the predicted and reference answers. We normalize the predicted and ground truth answers following the implementation of \citet{fang2025kirag}.

\textbf{Implementation details.} We implement our PAIRS framework using Qwen2.5-7B-Instruct \cite{team2024qwen2} as the backbone LLM for answer and pseudo-context generation. To ensure deterministic outputs and eliminate variability due to random sampling, we set the temperature to 0.0 during decoding \cite{kimsure}. For dense retrieval, we employ the bge-large-en-v1.5 embedding model \cite{bge_embedding} with a hidden dimension of 768, using inner product (IP) as the similarity metric. In the dual-path retrieval step, we retrieve the top 5 most relevant documents independently for both the query and the generated pseudo-context, resulting in a combined candidate pool of 10 documents ($2n = 10$). For reranking, we adopt the bge-reranker-base model \cite{bge_embedding} to reorder the top-10 documents retrieved by the query-only baseline. Across all experiments, we pass the top 3 documents to the LLM for final answer generation, except \emph{No Retrieval}. All remaining experimental configurations of baselines follow the implementations reported in their respective original papers. We present the prompts used in this study in the appendix.

\subsection{Main results}

Tab. \ref{tab: results} summarizes the performance of PAIRS and its variants against a range of baselines on six QA datasets. We report both EM and F1 scores, along with the retriever activation ratio (\emph{RA ratio}) of PAIRS across datasets. Overall, the proposed PAIRS framework consistently outperforms baseline methods. Specifically, (1) \textbf{PAIRS} surpasses all prior baselines---including Standard RAG, HyDE, Q2D, and CoT---with an average of 1.1\% EM and 1.0\% F1 score improvement while maintaining a retriever activation ratio of only 75.5\%. This demonstrates PAIRS’s ability to enhance both accuracy and efficiency by bypassing unnecessary retrieval for simple queries. (2) \textbf{DPR-AIS}, which integrates dual-path retrieval and adaptive information selection for all queries, further improves performance, achieving the highest average EM (35.60\%) and F1 score (36.55\%) among non-reranking models. The consistent gains across various datasets highlight the robustness of DPR-AIS in both open-domain and multi-hop QA scenarios. (3) \textbf{DPR-AIS-rerank} extends these gains even further by incorporating a reranker, respectively improving the EM and F1 score by 2.3\% and 2.5\% on average. Also, it improves over the Rerank baseline by 1.7\% EM and 1.5\% F1 score on average, reaching state-of-the-art results in 5 out of 6 datasets. These improvements validate the compatibility and synergy between DPR-AIS and post-retrieval reranking mechanisms. (4) Notably, on datasets like WebQuestions, where queries contain sparse signals and retrieval noise is common, PAIRS achieves higher performance than DPR-AIS, illustrating that adaptive retrieval can improve generation accuracy by skipping unhelpful or distracting documents.

\begin{table*}[!ht]
\centering
%\resizebox{.95\columnwidth}{!}{
\begin{tabular}{l c c c c c c c c c c}
\toprule
    \multirow{2}{*}{Method} & \multicolumn{2}{c}{HotpotQA} & \multicolumn{2}{c}{NaturalQA} & \multicolumn{2}{c}{SQuAQ} & \multicolumn{2}{c}{TriviaQA} & \multicolumn{2}{c}{Average}\\
        & EM & F1 & EM & F1 & EM & F1 & EM & F1 & EM & F1 \\
\midrule
    \rowcolor{lightgray!30}
    DPR-AIS & \textbf{36.61} & \textbf{47.02} &  \textbf{37.42} & \textbf{38.95} & \underline{30.00} & \underline{35.87} & \textbf{59.09} & \textbf{33.32} & \textbf{40.78} & \textbf{38.79} \\
    
    w/o DPR (w/ $q$)           & \underline{36.14} & \underline{46.36} & 36.21 & 37.92 & \textbf{30.10} & \textbf{35.99} & \underline{58.08} & \underline{32.85} & \underline{40.13} & \underline{38.28}\\
    w/o DPR (w/ $p$)           & 35.04 & 44.90 & \underline{36.73} & \underline{38.09} & 28.34 & 34.17 & 57.77 & 32.40 & 39.47 & 37.39 \\
    w/o AIS (w/ 2$q$ + 1$p$)   & 33.94 & 43.49 & 31.99 & 33.90 & 26.11 & 31.43 & 53.05 & 29.72 & 36.27 & 34.64 \\
    w/o AIS (w/ 1$q$ + 2$p$)   & 28.55 & 37.38 & 28.73 & 31.53 & 22.00 & 26.76 & 48.76 & 27.38 & 32.01 & 30.76 \\
    %w/o $\theta$ (IP)       & 36.81 & 47.09 & 37.37 & 39.00 & 30.37 & 36.28 & 59.25 & 33.43 \\
\bottomrule  
\end{tabular}
\caption{Ablation experiments on four datasets, where w/o DPR (w/ $q$) and w/o DPR (w/ $q$) remove dual-path retrieval but only uses $q$ and $p$ to retrieve documents, respectively; and w/o AIS (w/ 2$q$ + 1$p$) and w/o AIS (w/ 1$q$ + 2$p$) use fixed ratios of retrieved documents instead of adaptive selection. \textbf{Bold} and \underline{underlined} values represent the best and second-best scores, respectively.}
\label{tab: ablation}
\end{table*}

\subsection{Ablation study}

To better understand the contributions of different components within the DPR-AIS framework, we conduct a series of ablation studies across four representative QA datasets. The results are summarized in Tab. \ref{tab: ablation}; the complete DPR-AIS model consistently achieves the best performance across all four evaluated datasets, highlighting the effectiveness of our design.

Specifically, to assess the role of DPR, we examine two variants where one retrieval path is removed: \textit{w/o DPR (w/ $q$)} and \textit{w/o DPR (w/ $p$)}. In these settings, the system retrieves the top 10 documents using only the query $q$ or the generated pseudo-context $p$, respectively, and then applies AIS to select the final top 3 documents for generation. We find that removing either path leads to a performance drop. Notably, the drop is larger when using only $p$ (\textit{w/o DPR (w/ $p$)}), suggesting that the LLM-generated pseudo-context is not sufficient on its own. On the other hand, when relying only on the query (\textit{w/o DPR (w/ $q$)}), the performance remains relatively close to the full DPR-AIS model, indicating the robustness of query--based retrieval and its dominance in guiding relevance.

We then assess the effectiveness of the Adaptive Information Selection (AIS) module by replacing it with fixed document selection heuristics. In \textit{w/o AIS (w/ 2$q$ + 1$p$)} and \textit{w/o AIS (w/ 1$q$ + 2$p$)}, we retrieve documents using fixed proportions (e.g., two from query and one from pseudo-context, or vice versa), bypassing the adaptive scoring mechanism. These variants perform significantly worse than DPR-AIS across all datasets, with particularly steep drops on SQuAQ and TriviaQA. This highlights the necessity of adaptively weighing relevance from both sources. Please refer to the appendix for a more detailed analysis of the relationship among the query, pseudo-context, and retrievals.

\subsection{Additional analysis}\label{sec: analysis}

\textbf{Dynamic weighting between query and pseudo-context.} To better understand the impact of weighting between the query and the LLM--generated pseudo-context in the AIS module, we randomly sampled 5,000 queries from the HotpotQA training set and generated pseudo-contexts for each query using the LLM. We then explored the correlation between $\alpha$ (calculated using Eq. \eqref{eq: alpha}) and $\theta_0$ (the angle between the query and the pseudo-context) to examine how the divergence between the query and the generated context affects their relative importance (please refer to the appendix for details). We conducted a simple linear regression and obtained the relationship, expressed as:
\begin{equation}
    \alpha = 0.058 \theta_0 + 0.455.
\end{equation}
This result suggests that as the semantic gap between the query and pseudo-context increases, the optimal strategy is to place more emphasis on the original query during document selection. We then integrated this dynamic weighting mechanism into the AIS module and evaluated its effectiveness. As shown in Tab. \ref{tab: dynamic}, the dynamically weighted variant (DPR-AIS-dynamic) slightly improves over the fixed-weight version, achieving higher EM and F1 scores on HotpotQA. This result highlights the adaptability of AIS and the benefit of learning to balance information sources on a per-sample basis. In future work, more advanced strategies such as learned weighting mechanisms or neural attention over context sources could be explored to further enhance adaptive retrieval and selection.

\begin{table}[!ht]
\centering
%\resizebox{.95\columnwidth}{!}{
\begin{tabular}{l c c}
\toprule
    \multirow{2}{*}{Method} & \multicolumn{2}{c}{HotpotQA} \\
        & EM & F1 \\
\midrule
    DPR-AIS & 36.61 & 47.02 \\
    DPR-AIS-dynamic & \textbf{36.81} & \textbf{47.03} \\
\bottomrule  
\end{tabular}
\caption{Experimental results on HotpotQA with and without considering dynamic weighting between the query and LLM--generated pseudo-context.}
\label{tab: dynamic}
\end{table}

\begin{figure}[!ht]
    \centering
    \includegraphics[width=0.8\linewidth]{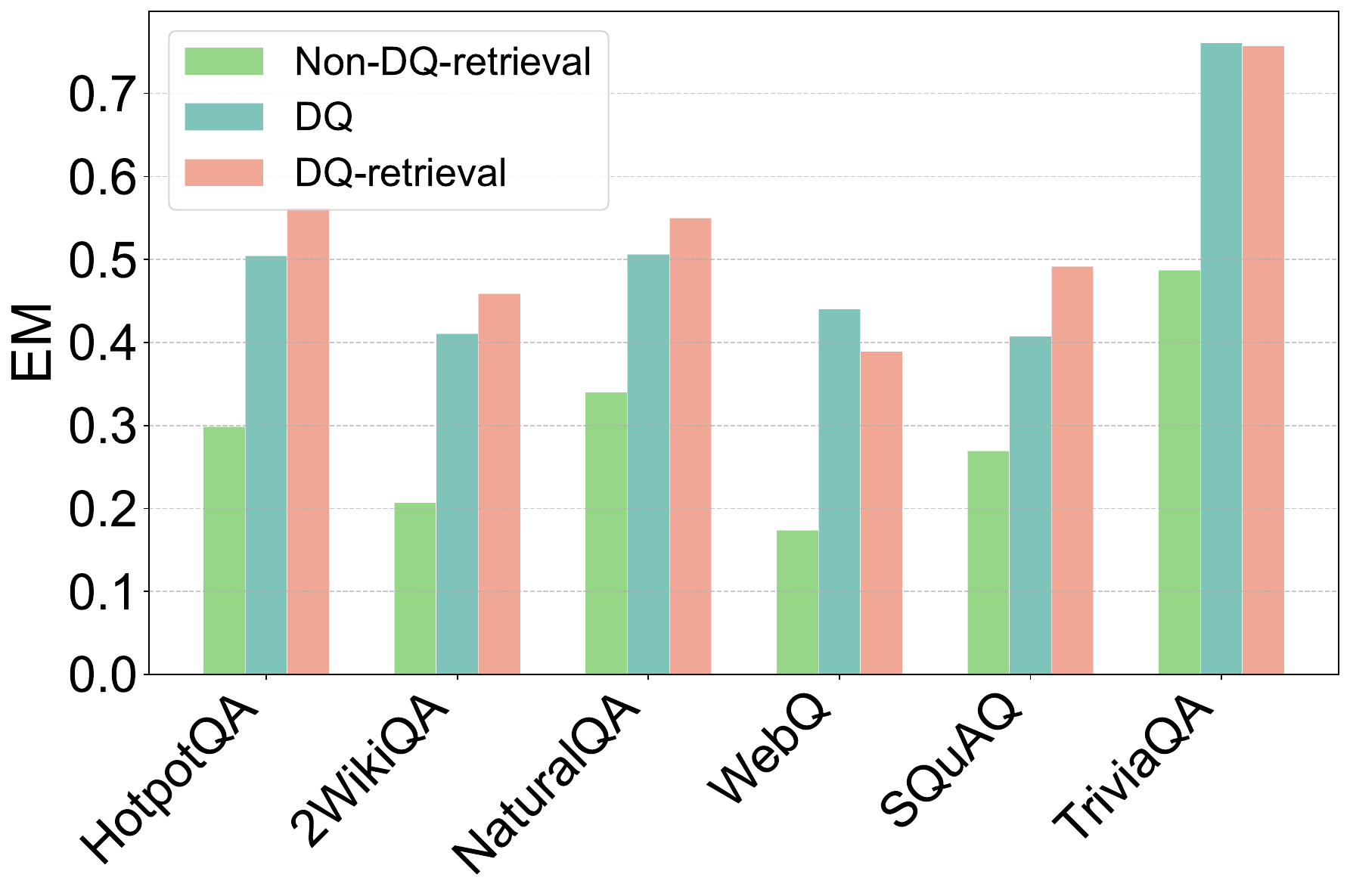}
    \caption{EM scores of different queries across six datasets}
    \label{fig: DQ_EM}
\end{figure}

\begin{table*}[!htbp]
\centering
\resizebox{2.1\columnwidth}{!}{
\begin{tabular}{l c c c c c c c c c c c c c c}
\toprule
    \multirow{2}{*}{Method} & \multicolumn{2}{c}{HotpotQA} & \multicolumn{2}{c}{2WikiQA} & \multicolumn{2}{c}{NaturalQA} & \multicolumn{2}{c}{WebQuestions} & \multicolumn{2}{c}{SQuAQ} & \multicolumn{2}{c}{TriviaQA} & \multicolumn{2}{c}{Average}\\
        & EM & F1 & EM & F1 & EM & F1 & EM & F1 & EM & F1 & EM & F1 & EM & F1\\
\midrule

    \rowcolor{lightgray!30}
    PAIRS & 35.14 & 45.20 & 26.99 & 31.73 & 36.87 & 38.17 & \textbf{23.08} & \textbf{31.15} & 28.85 & 34.90 & \textbf{59.23} & 32.92 & 35.03 & 35.68 \\
    
    RA ratio & \multicolumn{2}{c}{(74.4\%)} & \multicolumn{2}{c}{(69.1\%)} & \multicolumn{2}{c}{(82.8\%)} & \multicolumn{2}{c}{(78.6\%)} & \multicolumn{2}{c}{(86.3\%)} & \multicolumn{2}{c}{(61.6\%)} & \multicolumn{2}{c}{(75.5\%)}\\

    \rowcolor{lightgray!30}
    Exclude\_num & \textbf{35.30} & \textbf{45.38} & \textbf{27.00} & \textbf{31.75} & \textbf{37.04} & \textbf{38.33} & 22.98 & 31.13 & \textbf{29.22} & \textbf{35.28} & 59.18 & \textbf{32.93} & \textbf{35.12} & \textbf{35.80}\\
    
    RA ratio & \multicolumn{2}{c}{(76.1\%)} & \multicolumn{2}{c}{(69.8\%)} & \multicolumn{2}{c}{(87.9\%)} & \multicolumn{2}{c}{(80.7\%)} & \multicolumn{2}{c}{(89.9\%)} & \multicolumn{2}{c}{(64.2\%)} & \multicolumn{2}{c}{(78.1\%)}\\
    
\bottomrule
\end{tabular}}
\caption{Effects of LLM hallucinations on directly generated answers using parametric knowledge.}
\label{tab: LLM_hall}
\end{table*}

\begin{figure}
    \centering
    \includegraphics[width=1.0\linewidth]{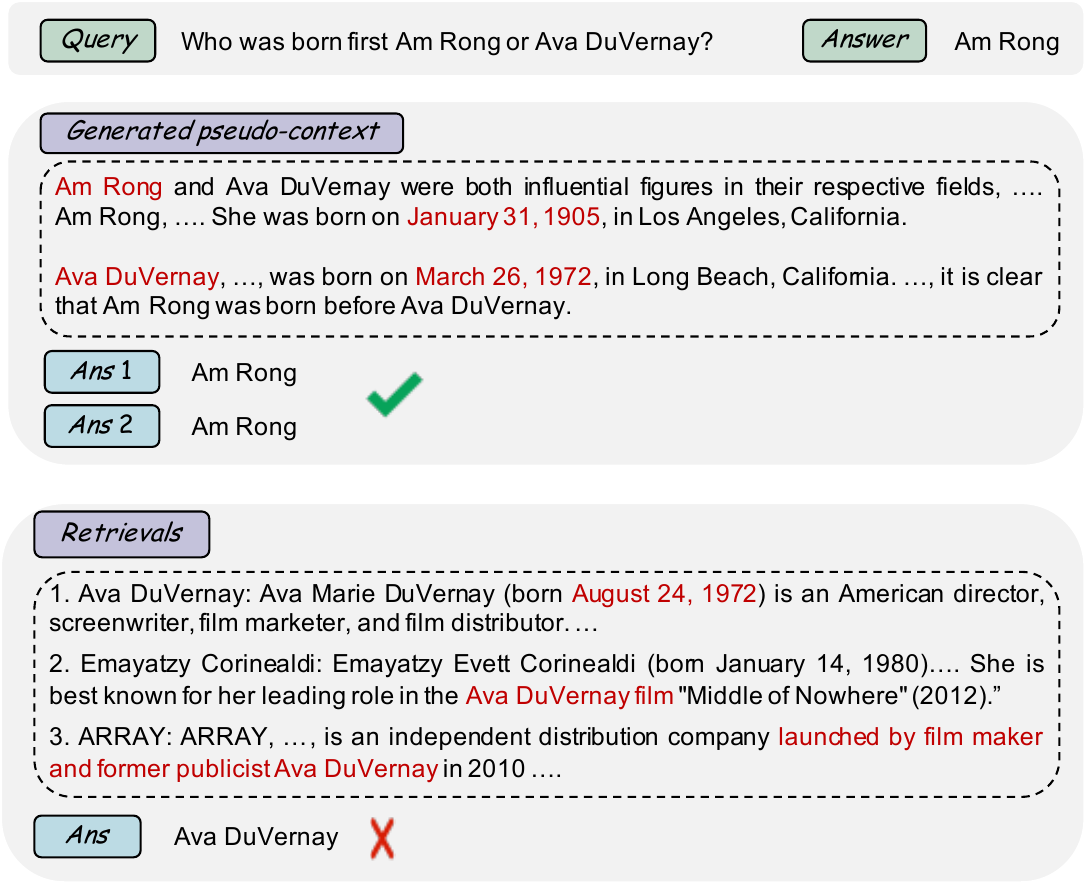}
    \caption{Qualitative comparison between PARIS and standard RAG.}
    \label{fig: case_study}
\end{figure}

\textbf{Parametric knowledge can generate accurate answers.} To further investigate the LLM’s ability to generate accurate answers without relying on external retrieval, we categorize the test queries into three groups: (1) queries that are answered by the LLM directly, referred to as Directly Answerable Queries (\textbf{DQ}); (2) the remaining queries that require retrievals, referred to as \textbf{Non-DQ-retrieval}; and (3) a control setting where the same DQs are answered using retrieval, denoted as \textbf{DQ-retrieval}. Fig. \ref{fig: DQ_EM} presents the exact match (EM) results across six datasets (with corresponding F1 scores and other quantitative results provided in the appendix). We observe that the EM scores achieved by the LLM on DQ are consistently high, and in some cases, even surpass those of DQ-retrieval. Both DQ and DQ-retrieval substantially outperform Non-DQ-retrieval, indicating that DQs are generally simpler or factually grounded queries that fall within the LLM’s pre-trained knowledge base. These results indicate that our dual-path generation mechanism not only allows accurate answer generation directly but also supports selective retrieval when needed---offering a more efficient and adaptive RAG framework.

\textbf{Effects of LLM hallucinations.} To assess the impact of hallucinations when large language models (LLMs) generate answers without retrieval, we conduct a controlled experiment based on a simple heuristic: if a generated answer contains numeric values, it is more likely to be affected by hallucination. This is because LLMs are generally less reliable when producing precise facts such as numbers, dates, or counts from parametric memory alone \cite{ji2023survey, singh2025agentic}. We filter out all directly answered queries (DQs) whose generated answers contain numbers, and we then rerun our DPR-AIS for these queries (referred to \textit{Exclude\_num}). The results are reported in Tab. \ref{tab: LLM_hall}. Overall, excluding numeric DQs results in slightly improved performance. The average exact match (EM) increases from 35.03 to 35.12, and the average F1 score improves from 35.68 to 35.80. While these gains are modest, they come with an increase in the retriever activation (RA) ratio---from 75.5\% to 78.1\%. These findings suggest that hallucinations from LLMs, especially when dealing with factual numeric content, can indeed affect answer quality in retrieval-free settings. However, their overall impact remains limited in our setting, as the performance gain is marginal. This reinforces the strength of PAIRS: it naturally handles simple queries with direct parametric answers with trivial effects of LLM hallucinations.

\textbf{Case study.} Fig. \ref{fig: case_study} demonstrates a representative example. The query is a simple factual comparison that lies well within the LLM's parametric knowledge, and as shown in the generated pseudo-context, the model accurately generates birth dates and correctly answers \textit{Am Rong}. However, the standard RAG generates an incorrect answer because the retrievals present \textit{Ava DuVernay}'s birth year (1972) only, lacking complete evidence on \textit{Am Rong}. This example demonstrates that LLMs can generate accurate answers to simple queries using their internal parametric knowledge, while noisy or incomplete retrievals---especially those missing critical comparative context---can mislead the final generation. We present more cases in the appendix.

%%%%%%%%%%%%%%%%%%%%%%%%%%%%%%% Conclusion %%%%%%%%%%%%%%%%%%%%%%%%%%%%%%%
\section{Conclusion}

In this paper, we propose PAIRS, a simple yet effective framework that adaptively integrates parametric and retrieved knowledge for question answering. By verifying the LLM’s parametric response and selectively activating retrieval with dual-path signals and adaptive selection, PAIRS improves both accuracy and efficiency. Experiments across multiple QA benchmarks demonstrate that PAIRS consistently outperforms existing baselines and can be easily applied to real-world RAG systems.

%%%%%%%%%%%%%%%%%%%%%%%%%%%%%%% Reference %%%%%%%%%%%%%%%%%%%%%%%%%%%%%%%
\newpage
\bibliography{refs}

%%%%%%%%%%%%%%%%%%%%%%%%%%%%%%% Appendix %%%%%%%%%%%%%%%%%%%%%%%%%%%%%%%

\newpage

\appendix

\section{Notation List}

The main notations and abbreviations used in this paper are listed in Tab. \ref{tab: notation}.

\begin{table}[!ht]
\centering
\resizebox{0.99\columnwidth}{!}{
\begin{tabular}{l l}
\toprule
    Notation &  Explanation\\
\midrule
    $D$           & Chunks of Documents \\
    $D_k$         & Retrieved $k$ chunks \\
    $f$           & Embedding model \\
    $q$           & Query \\
    $\mathbf{q}$  & Query embedding \\
    $d$           & A chunk \\
    $\mathbf{d}$  & Chunk embedding \\
    $\mathcal{R}$ & Retriever \\
    $\mathcal{P}$ & Prompt \\
    $\mathcal{G}$ & LLM generator \\
    $\hat{a}$     & Generated answer \\
    $p$           & LLM--generated pseudo-context \\
    $\mathbf{p}$  & Pseudo-context embedding \\
    $\theta_0$    & The angle between $\mathbf{q}$ and $\mathbf{p}$ \\
    $\theta_1$    & The angle between $\mathbf{q}$ and $\mathbf{d}$ \\
    $\theta_2$    & The angle between $\mathbf{p}$ and $\mathbf{d}$ \\
    $s(d)$        & Score of $d$ \\
    $\alpha$      & Importance factor of $\theta_1$ \\
    $\mathcal{K}$ & Reranker \\
\midrule
    Abbreviation &  Explanation \\
\midrule
    PAIRS & Parametric--verified adaptive information \\
          & retrieval and selection \\
    DPR   & Dual-path retrieval \\
    AIS   & Adaptive information selction \\
    IP    & Inner product \\
    EM    & Exact match \\
    RA ratio & Retriever activation ratio \\
    
\bottomrule  
\end{tabular}}
\caption{Explanation of main notations and abbreviations used in this study.}
\label{tab: notation}
\end{table}

\section{Retrieval Analysis}

In this section, we further analyze the properties of retrievals using the query or LLM--generated pseudo-context. Specifically, we first compare the documents retrieved using the query with the ground-truth (GT) documents. We found that the retrieved documents may be significantly different from the GT documents. In addition, we analyze the spatial distribution of query--based, pseudo-context--based, and GT documents. We discovered that the documents retrieved using the pseudo-context may compensate for this shortcoming.

\subsection{Similarity between query and GT documents}

\begin{figure}[!ht]
    \centering
    \includegraphics[width=1.0\linewidth]{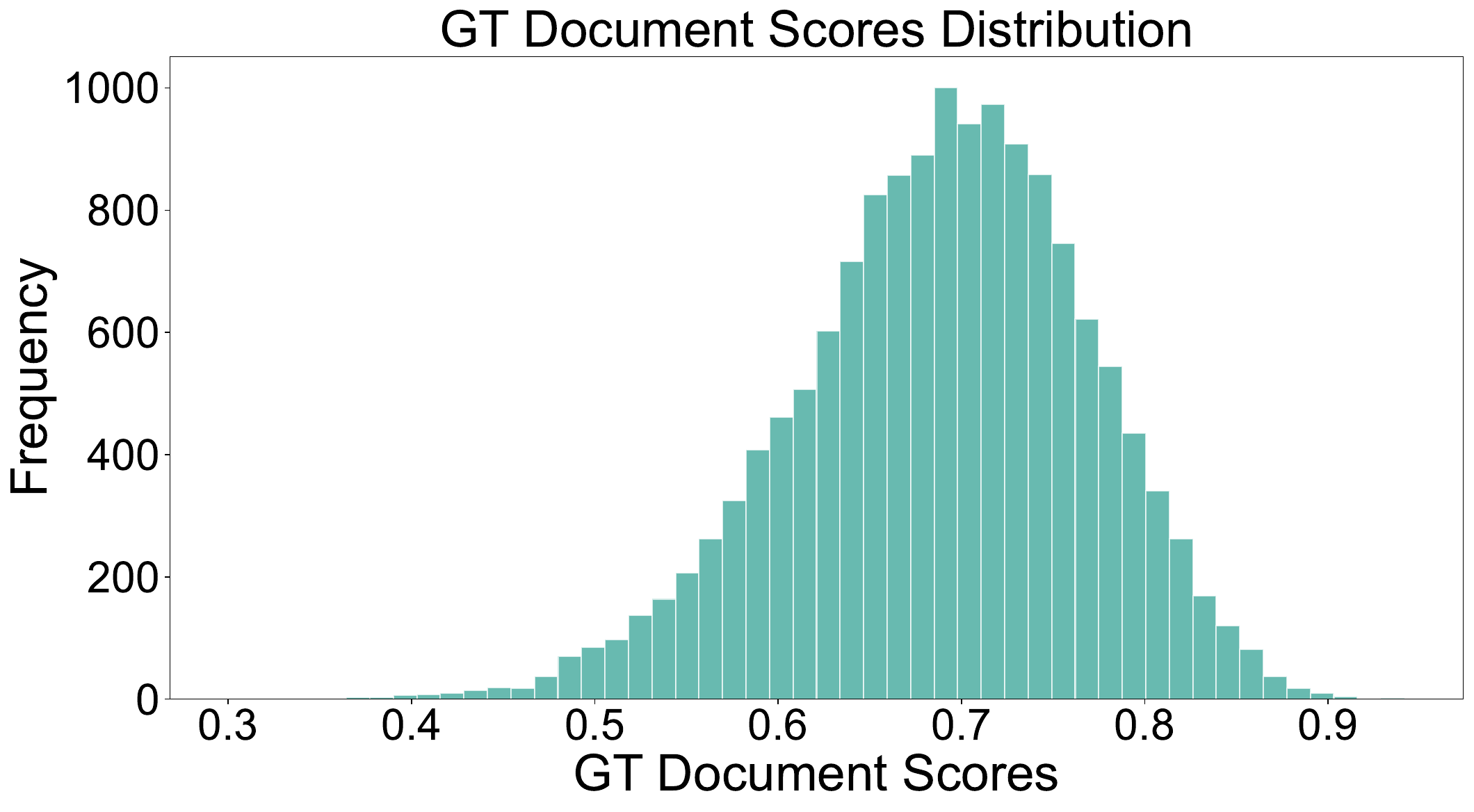}
    \caption{Similarity scores between queries and ground truth documents.}
    \label{fig: GT_doc_score}
\end{figure}

To figure out how the query--based documents differ from the GT documents, we first calculated the similarity scores between the ground-truth(GT) documents and the queries in the HotpotQA dataset. As shown in Fig. \ref{fig: GT_doc_score}, the similarity scores of the most GT documents are below 0.8, indicating that the query may differ from the GT documents, which can lead to irrelevant retrievals.

\begin{figure}[!ht]
    \centering
    \includegraphics[width=1.0\linewidth]{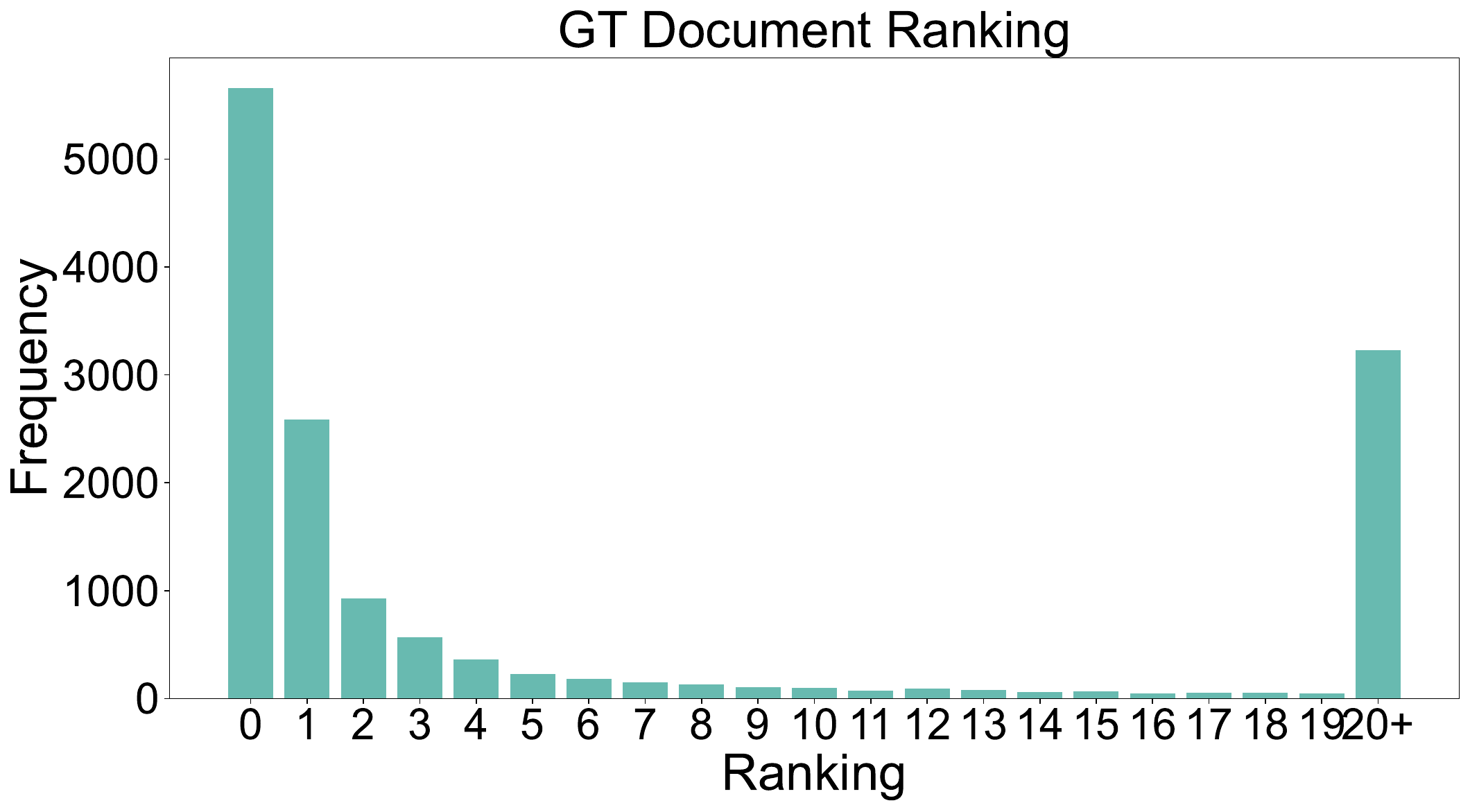}
    \caption{Ranking of ground truth documents in query--based retrievals.}
    \label{fig: GT_doc_orders}
\end{figure}

Furthermore, we ranked the GT documents in the query--based retrievals. As shown in Fig. \ref{fig: GT_doc_orders}, a significant proportion of the GT documents rank 20+ in the documents retrieved using the query. This further demonstrates that using the query to retrieve documents only can lead to irrelevant results, which in turn result in inaccurate final answers.

\subsection{Spatial distribution of query, context, and GT documents}

\begin{figure}[!ht]
    \centering
    \includegraphics[width=0.7\linewidth]{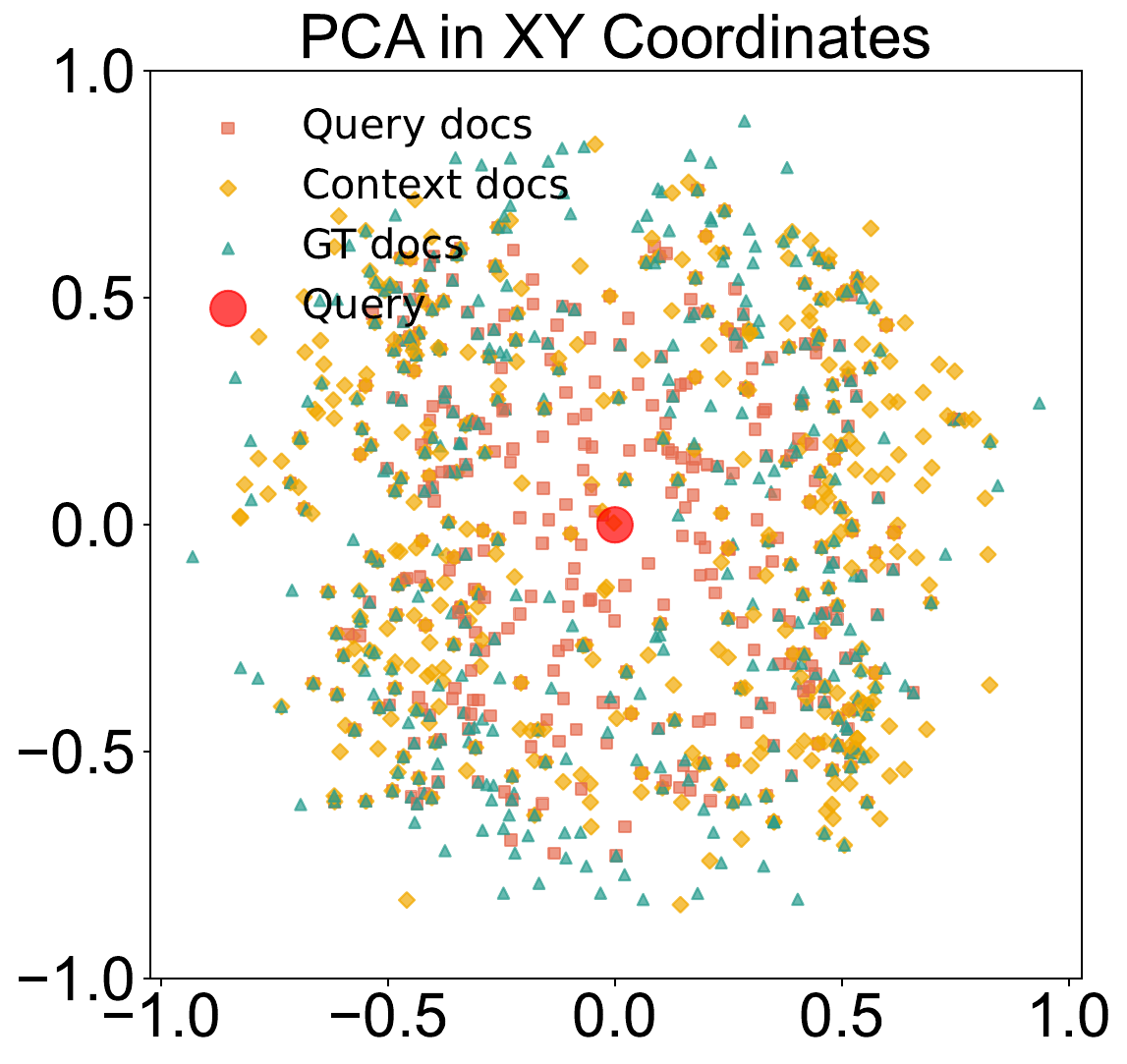}
    \caption{Relationship between query, context, and GT documents in the XY coordinate.}
    \label{fig: pca_xy}
\end{figure}

\begin{figure}[!ht]
    \centering
    \includegraphics[width=0.7\linewidth]{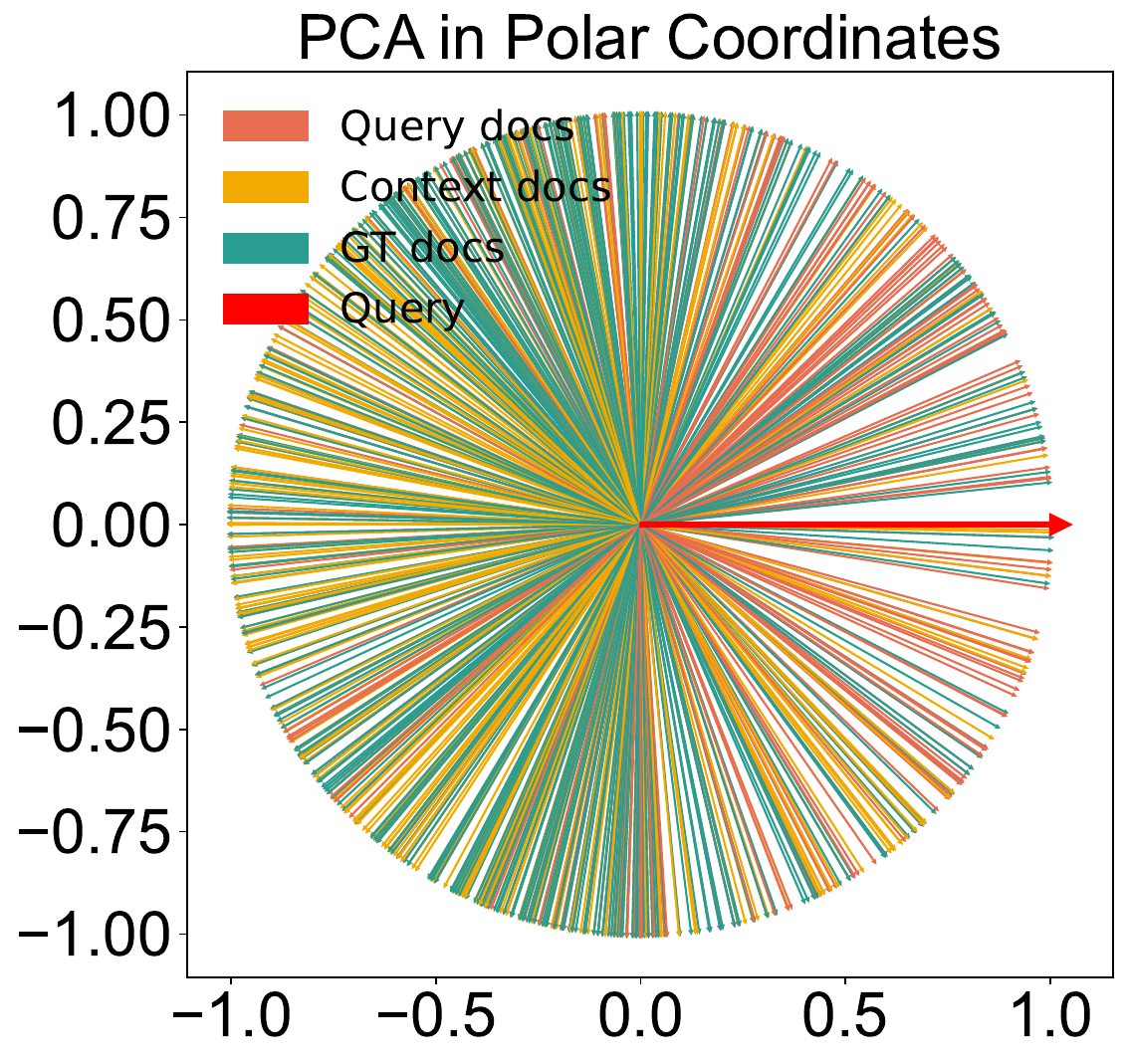}
    \caption{Relationship between query, context, and GT documents in the polar coordinate.}
    \label{fig: pca_polar}
\end{figure}

To figure out the relationship between the query, context, and GT documents, we projected these documents into the 2D space. Figs. \ref{fig: pca_xy} and \ref{fig: pca_polar} illustrate their relationship in xy and polar coordinates, respectively. Remind that we only visualize 200 samples, and we set the origin of the coordinates as the query. Most of the query documents are located besides the origin (i.e., the query), while the GT documents are kind of far from the origin, demonstrating that the GT documents may differ from the query documents in the vector space. However, the context documents can compensate for this gap as their distribution is similar to that of the GT documents when they are far away from the origin. These results further demonstrate that the LLM--generated pseudo-context can compensate for the sparse queries, enhancing the performance of RAG systems.

\section{Prompt Template}

\begin{figure}[!ht]
    \centering
    \includegraphics[width=1.0\linewidth]{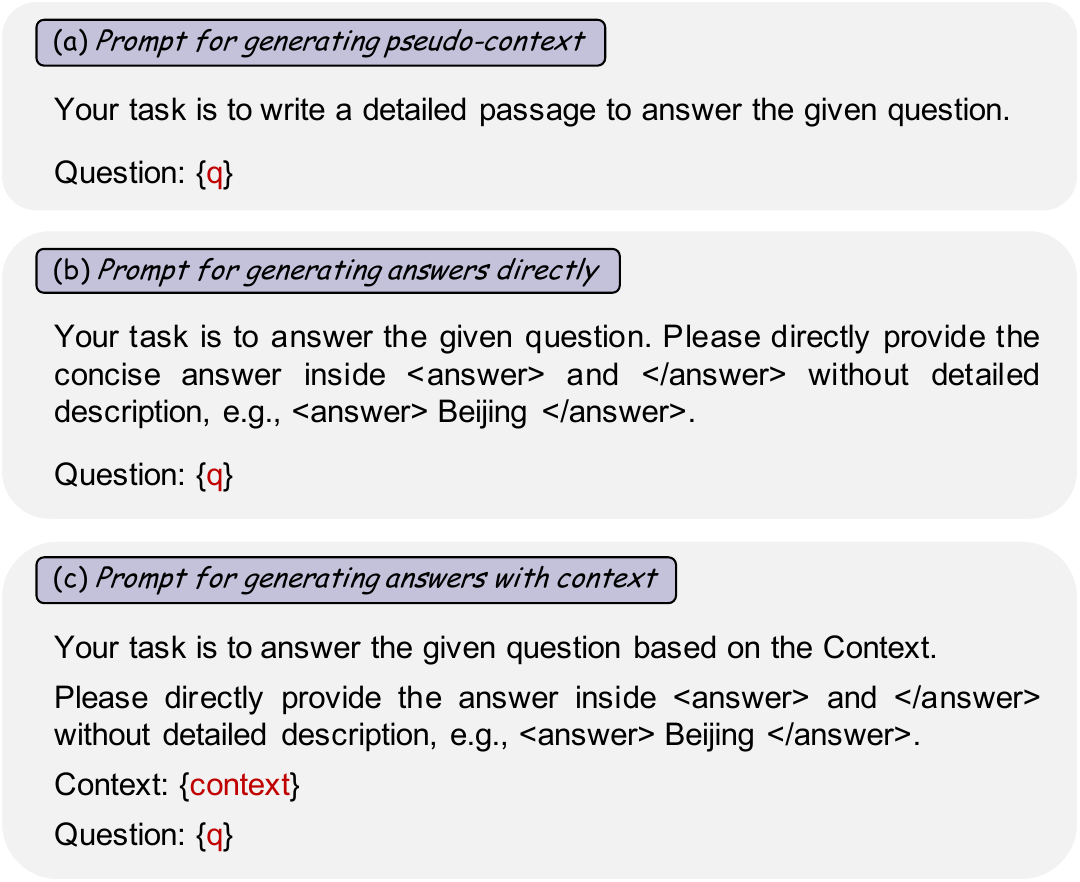}
    \caption{Prompt templates used in this paper, including (a) generating pseudo-context, (b) generating answers directly, and (c) generating answers based on context. \color{red}{q} \color{black}{and} \color{red}{context} \color{black} denote the query and generated or retrieved context, respectively.}
    \label{fig: prompt}
\end{figure}

The prompt templates used in this paper for generating pseudo-context or answers are presented in Fig. \ref{fig: prompt}.

\section{Details of Dynamic Weighting}

\begin{figure}[!ht]
\centering
\includegraphics[width=0.95\columnwidth]{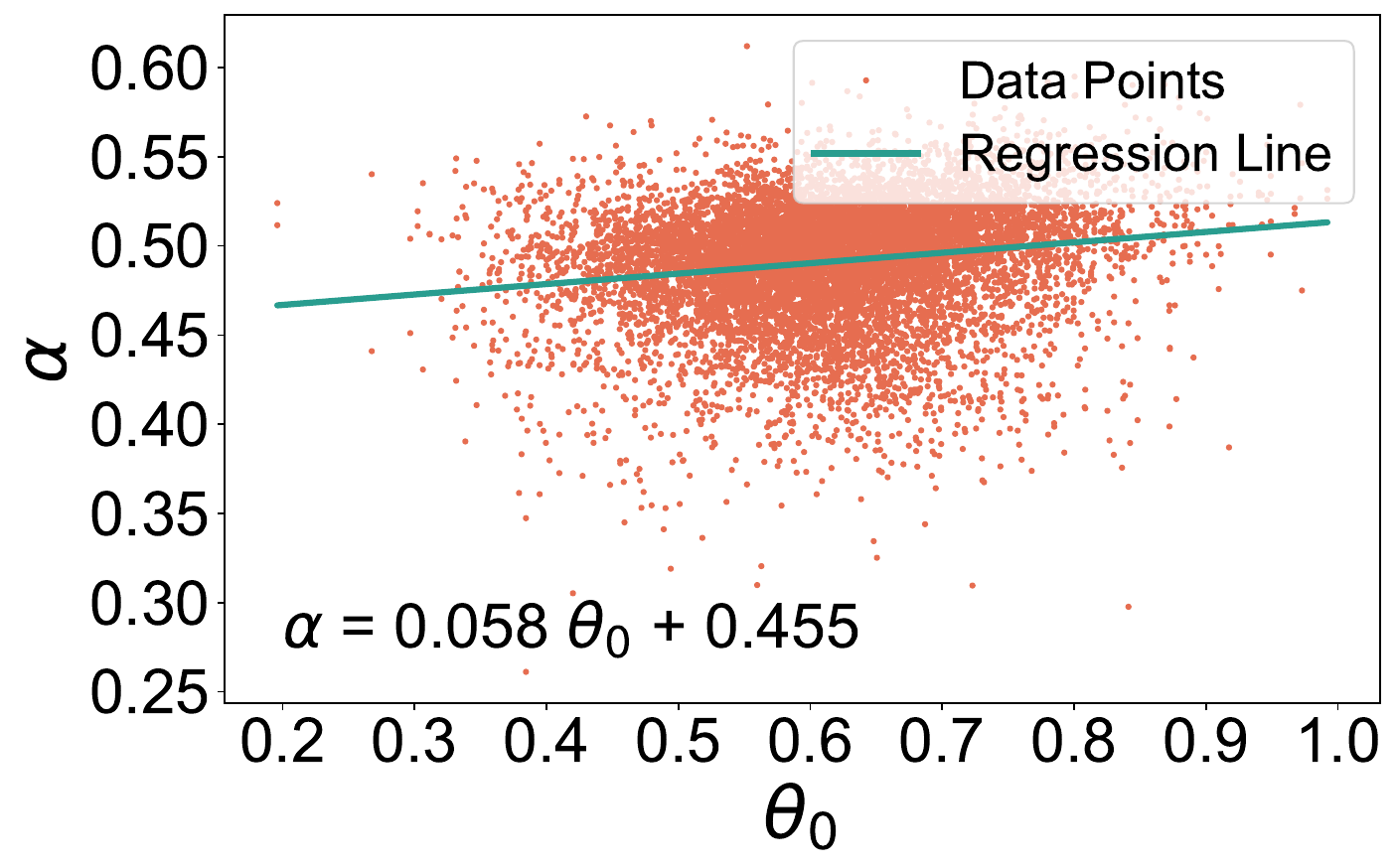}
\caption{Relationship between the importance factor $\alpha$ and the angle between the embeddings of the query and LLM--generated pseudo-context. A linear relationship can be obtained through regression as $\alpha = 0.058\theta_0 + 0.455$.}
\label{fig: dynamic_weight}
\end{figure}

To analyze the relationship between queries, pseudo-contexts, and ground-truth documents, we randomly sampled 5,000 queries from the HotpotQA training set and generated pseudo-contexts for each using an LLM. For every query, we measured three key angles in the embedding space: $\theta_0$, $\theta_1$, and $\theta_2$. Using these angles, we derived the optimal weighting factor $\alpha = \frac{\theta_2}{\theta_1 + \theta_2}$. We then investigated how $\alpha$ correlates with $\theta_0$ to assess the influence of semantic divergence between queries and pseudo-contexts on their relative importance in document selection. As illustrated in Fig. \ref{fig: dynamic_weight}, a linear regression analysis revealed a clear positive relationship, which indicates that as the semantic gap ($\theta_0$) widens, the optimal strategy increasingly prioritizes the original query over the generated pseudo-context.

\begin{table*}[!ht]
\centering
\resizebox{2.1\columnwidth}{!}{
\begin{tabular}{l c c c c c c c c c c c c c c}
\toprule
    \multirow{2}{*}{Method} & \multicolumn{2}{c}{HotpotQA} & \multicolumn{2}{c}{2WikiQA} & \multicolumn{2}{c}{NaturalQA} & \multicolumn{2}{c}{WebQuestions} & \multicolumn{2}{c}{SQuAQ} & \multicolumn{2}{c}{TriviaQA} & \multicolumn{2}{c}{Average}\\
        & EM & F1 & EM & F1 & EM & F1 & EM & F1 & EM & F1 & EM & F1 & EM & F1\\
\midrule
Standard RAG & \underline{36.49} & \underline{46.84} & \textbf{30.15} & \textbf{35.04} & 35.46 & 37.48 & 21.31 & 29.10 & \underline{31.29} & 37.22 & 57.36 & 32.49 & 35.34 & 36.36\\
    
    \rowcolor{lightgray!30}
    PAIRS & 35.07 & 45.10 & 28.09 & 32.89 & \underline{36.65} & \underline{38.63} & \textbf{23.67} & \textbf{31.93} & 31.03 & \underline{37.35} & \underline{60.70} & \underline{33.68} & \underline{35.87} & \underline{36.60}\\
    
    RA ratio & \multicolumn{2}{c}{(74.4\%)} & \multicolumn{2}{c}{(69.1\%)} & \multicolumn{2}{c}{(82.8\%)} & \multicolumn{2}{c}{(78.6\%)} & \multicolumn{2}{c}{(86.3\%)} & \multicolumn{2}{c}{(61.6\%)} & \multicolumn{2}{c}{(75.5\%)}\\
    
    \rowcolor{lightgray!30}
    DPR-AIS & \textbf{36.79} & \textbf{47.06} & \underline{29.71} & \underline{34.70} & \textbf{37.06} & \textbf{39.22} & \underline{22.54} & \underline{31.29} & \textbf{32.59} & \textbf{38.70} & \textbf{61.20} & \textbf{34.29} & \textbf{36.65} & \textbf{37.54}\\

\bottomrule
\end{tabular}}
\caption{Results across six QA datasets with top-5 documents. \textbf{RA ratio} indicates the proportion of queries for which the retriever was activated by PAIRS. \textbf{Bold} and \underline{underlined} values represent the best and second-best scores, respectively.}
\label{tab: results_top5}
\end{table*}

\section{Additional Results on Parametric Generation}

\begin{figure}[!ht]
    \centering
    \includegraphics[width=0.95\linewidth]{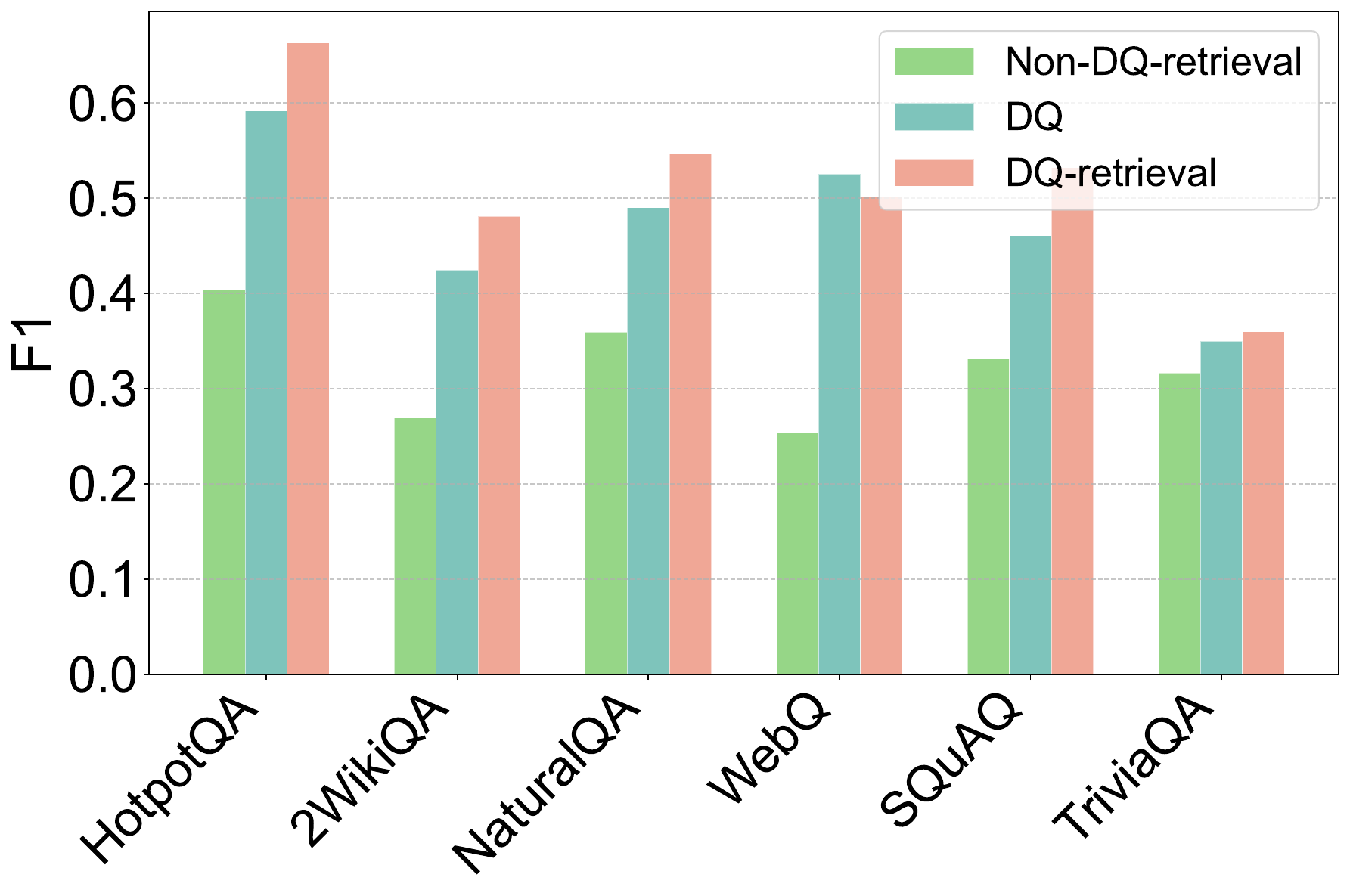}
    \caption{F1 scores of different queries across six datasets}
    \label{fig: DQ_F1}
\end{figure}

Fig. \ref{fig: DQ_F1} presents the F1 scores across six datasets. Similar to the EM scores, the F1 scores achieved on both DQ and DQ-retrieval substantially outperform those achieved on Non-DQ-retrieval, indicating that the parametric knowledge can generate accurate answers to simple queries. Tab. \ref{tab: DQ} presents the averaged exact match (EM) and F1 scores across six datasets. We observe that Non-DQ-retrieval queries yield the lowest performance, with an average EM of 29.61 and F1 of 32.22. In contrast, DQ queries---those answered without retrieval---achieve a significantly higher EM of 50.50 and F1 of 47.37. Interestingly, when retrieval is enforced for these DQ queries (i.e., DQ-retrieval), the performance only slightly improves to an EM of 53.49 and F1 of 51.39. These results indicate that the proposed dual-path generation and verification mechanism can achieve comparable accuracy to that based on retrievals on relatively simple queries.

\begin{table}[!ht]
\centering
%\resizebox{.95\columnwidth}{!}{
\begin{tabular}{l c c}
\toprule
    \multirow{2}{*}{Queries} & \multicolumn{2}{c}{Average} \\
        & EM & F1 \\
\midrule
    Non-DQ-retrieval & 29.61 &  32.22 \\
    DQ &  \underline{50.50} & \underline{47.37} \\
    DQ-retrieval & \textbf{53.49} & \textbf{51.39} \\
\bottomrule  
\end{tabular}
\caption{Average experimental results of different queries across six datasets.}
\label{tab: DQ}
\end{table}

\section{Additional Experiments}

We also conducted experiments with top-5 documents, and the results are listed in Tab. \ref{tab: results_top5}. Please note that we only consider the standard RAG baseline and our proposed method because (1) the experimental results with top-3 documents have demonstrated that the standard RAG can be representative and achieve comparable accuracy with the other baselines, and (2) conducting experiments across all baselines can be time-consuming. The experimental results are consistent with those achieved with top-3 documents, where PAIRS can achieve higher accuracy with triggering only 75\% retrievals. Also, DPR-AIS can further improve the performance by selecting the top-5 documents for all queries. These results further validate the effectiveness of the proposed method, which can not only enhance the efficiency but also improve the performance of RAG systems.

\section{Additional Case Studies}

\begin{figure}[!ht]
    \centering
    \includegraphics[width=0.99\linewidth]{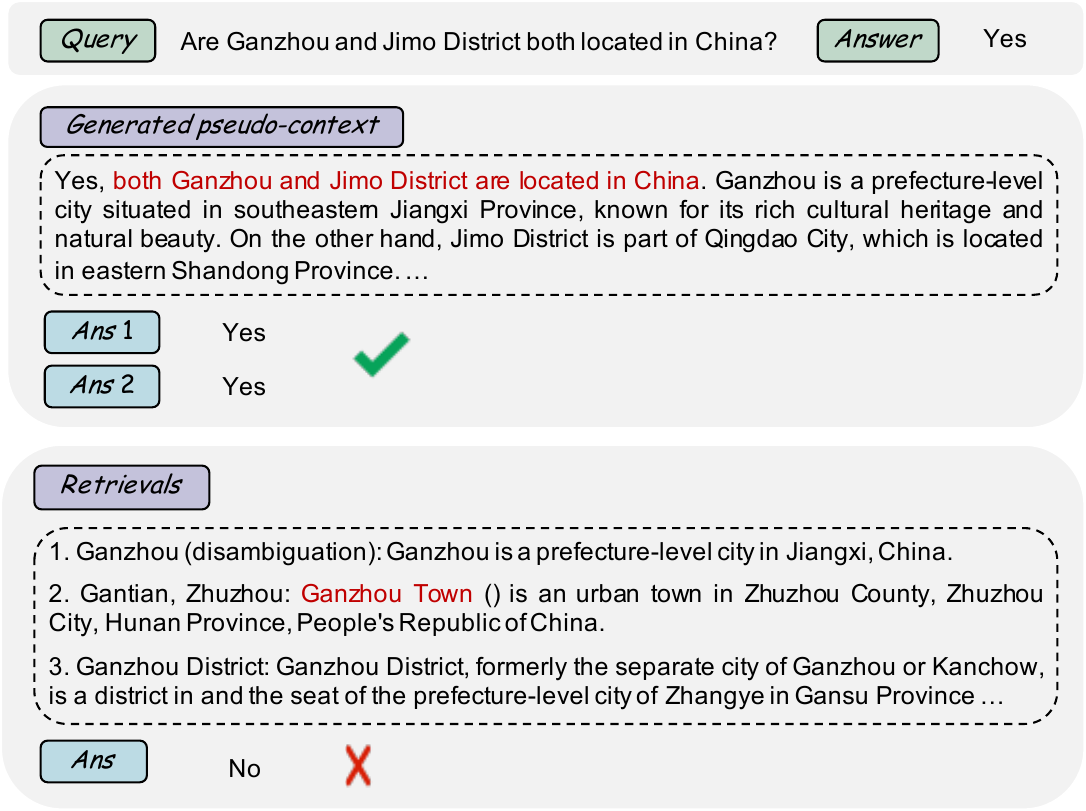}
    \caption{Comparison between dual-path generation mechanism and standard RAG.}
    \label{fig: case2}
\end{figure}

\begin{figure}[!ht]
    \centering
    \includegraphics[width=0.99\linewidth]{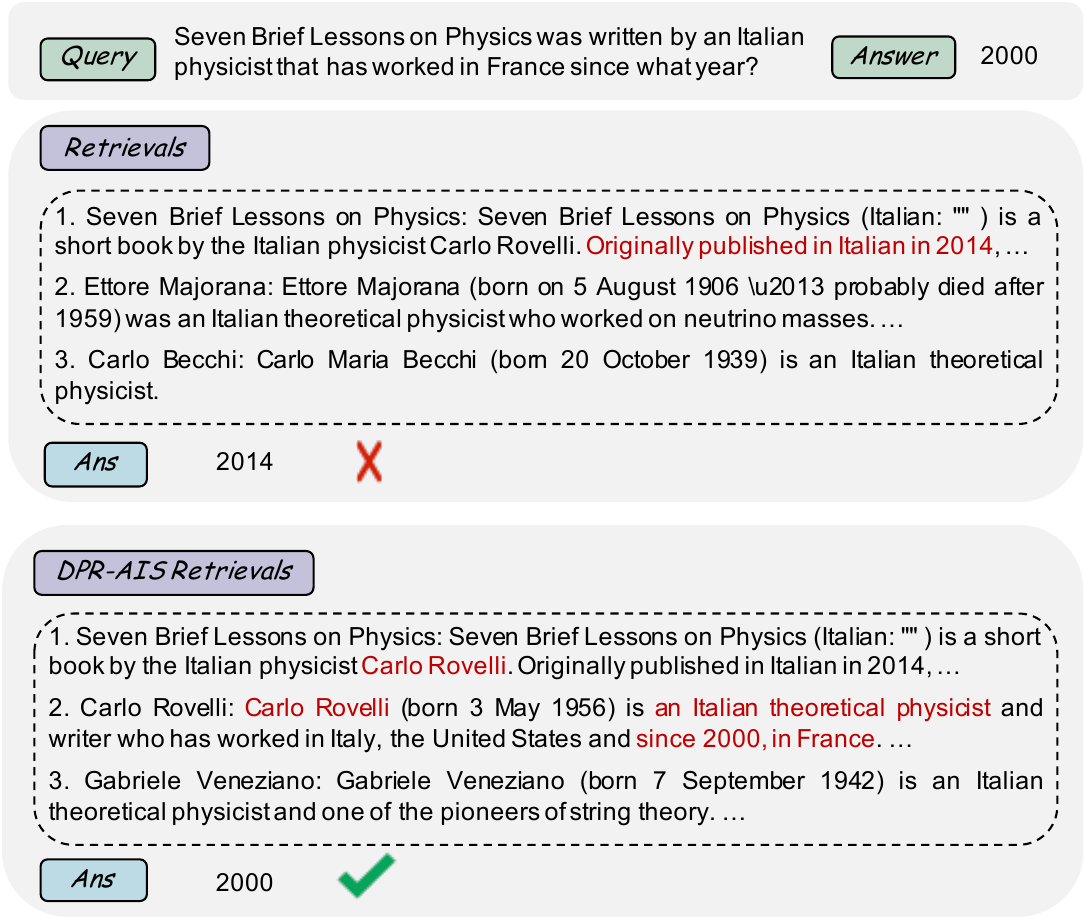}
    \caption{Comparison between standard RAG and DPR-AIS.}
    \label{fig: case3}
\end{figure}

In this section, we present additional case studies that demonstrate the effectiveness of our proposed method (PAIRS) and the standard RAG baseline. Fig. \ref{fig: case2} illustrates that PAIRS can generate accurate answers to queries that lie in the LLM's parametric domain, and the dual-path generation mechanism can effectively verify the final answer. However, irrelevant documents retrieved using the query can lead to an inaccurate answer.

Fig. \ref{fig: case3} compares the retrievals and answers achieved by standard RAG and DPR-AIS, respectively. Using the query to retrieve documents only can lead to similar but irrelevant results, which in turn results in a wrong final answer. However, our proposed dual-path retrieval mechanism can compensate for sparse queries and retrieve more relevant information, ultimately leading to accurate answers.

\end{document}